\documentclass{article}

\usepackage{arxiv}

\usepackage[utf8]{inputenc} 
\usepackage[T1]{fontenc}    
\usepackage{hyperref}       
\usepackage{url}            
\usepackage{booktabs}       
\usepackage{amsfonts}       
\usepackage{nicefrac}       
\usepackage{microtype}      
\usepackage{lipsum}
\usepackage{graphicx}
\graphicspath{ {./images/} }

\usepackage{microtype}
\usepackage{graphicx}
\usepackage{subfigure}
\usepackage{booktabs} 

\usepackage{amsmath}
\usepackage{bm}
\usepackage{amssymb}
\usepackage{amsfonts}
\usepackage{mathrsfs}
\usepackage{mathtools}

\usepackage{algorithm}
\usepackage{algorithmic}

\usepackage{natbib}

\usepackage{hyperref}

\title{Statistical Guarantees and Algorithmic Convergence Issues  \\
           of Variational Boosting}

\author{
 Biraj Subhra Guha \\
  PhD Candidate,
  Department of Statistics\\
  Texas A \& M University\\
   \And
 Anirban Bhattacharya \\
  Associate Professor,
  Department of Statistics\\
  Texas A \& M University\\
  \And
 Debdeep Pati \\
  Associate Professor,
  Department of Statistics\\
  Texas A \& M University\\
}

\begin{document}
\maketitle
\begin{abstract}
We provide statistical guarantees for Bayesian variational boosting by proposing a novel small bandwidth Gaussian mixture variational family. We employ a functional version of Frank-Wolfe optimization as our variational algorithm and study frequentist properties of the iterative boosting updates. Comparisons are drawn to the recent literature on boosting, describing how the choice of the variational family and the discrepancy measure affect both convergence and finite-sample statistical properties of the optimization routine. Specifically, we first demonstrate stochastic boundedness of the boosting iterates with respect to the data generating distribution. We next integrate this within our algorithm to provide an explicit convergence rate, ending with a result on the required number of boosting updates.
\end{abstract}


\section{INTRODUCTION}

Variational Bayes has gained popularity in recent years as an alternative to Markov chain Monte Carlo procedures to approximate analytically intractable posterior distributions; refer to \citet{blei2017variational} for a comprehensive overview. Variational inference formulates the problem of approximating the posterior as an optimization routine by minimizing a measure of discrepancy between probability densities in an approximating class and the posterior density. The variational solution refers to the closest member of the approximating class to the posterior, with closeness measured through divergences or metrics, usually Kullback--Leibler divergence. Other discrepancy measures for approximating the posterior have been studied, like the Wasserstein distance and R\'{e}nyi divergence in \cite{huggins2020validated}, Fisher distance in \cite{huggins2018practical} and Hellinger metric in \cite{campbell2019universal}.

The approximating class or the domain of optimization, commonly referred to as the variational family, plays a central role in these methods. It is chosen to strike a balance between computational tractability and approximation power. A richer, more flexible family allows better approximation of the posterior, while a simpler class of distributions facilitate calculations and computation speed. The Gaussian family is a popular example of a parametric variational family, where the optimization effectively takes place over a finite-dimensional parameter space. For a semi-parametric approach, one can use the popular mean-field family, which only assumes that the variational density factorizes over pre-specified sub-blocks of the parameter, with the factors otherwise unrestricted.

Statistical guarantees, frequentist validation as well as convergence issues focusing on mean-field appear in works like \citet{yang2017alpha}, \citet{pati2018statistical}, \citet{alquier2017concentration},  \citet{zhang2017convergence}, \citet{wang2018frequentist}, \citet{mukherjee2018mean}, \citet{yang2018mean} and \cite{huggins2020validated}. However, mean-field approximations can only hope to recover the center of the posterior and fails to capture posterior co-dependence, so need for more general families arise. Copula modelling has been used in \citet{tran2015copula} and \citet{han2016variational}, while implicit distribution families have been used in \citet{huszar2017variational}, \citet{han2016variational}, \citet{titsias2018unbiased}, \citet{yin2018semi}, \citet{molchanov2018doubly} and \citet{shi2017kernel}. Another recent approach to gain modelling flexibility is to use mixture distributions as variational families, which is the focus of this paper. \citet{wang2006convergence} is an early theoretical work on Gaussian mixtures as variational family, focusing on conjugate priors. As proposed in \citet{guo2016boosting} and \citet{miller2017variational}, the structure of mixture families naturally gives rise to the idea of variational boosting. This computation method iteratively builds a mixture distribution approximation to the posterior by adding simple, new components and re-weighting them. The components of the mixture can thus be considered weak learners in this boosting framework, which are averaged in a weighted, sequential fashion to produce a mixture, the strong learner.

Variational boosting offers better computational efficacy due to iterative fitting, while simultaneously improving approximation prowess owing to the more flexible mixture distribution class. \citet{guo2016boosting} modify the boosting method based on $L_2$--regularized variational objective. \citet{miller2017variational} incorporate covariance structure to modify the variational family. \citet{locatello2017boosting} provide some theoretical basis of this computational method using truncated densities as mixture components. However, their result is limited to compactly supported densities only, thus technically not including even Gaussian distributions. This idea is extended in \citet{locatello2018boosting} for black box variational inference; refer to \citet{ranganath2014black} for the original work on black box variational inference.  \citet{wang2016boosting} uses gradient boosting technique, and suffers from a drawback similar to \citet{locatello2017boosting}. \citet{campbell2019universal} note that the domain of mixture families does not allow Kullback--Leibler divergence to be sufficiently smooth, hence switches to Hellinger metric to provide algorithm and theoretical study for boosting. We address the above problem with the boosting method by providing a pathway to work with mixture families and simultaneously maintaining the use of Kullback--Leibler divergence.

Our contribution to variational boosting revolves around frequentist properties of the variational solution. We study this by proposing a small bandwidth Gaussian mixture variational family and using a functional version of the Frank--Wolfe algorithm (refer to \citet{frank1956algorithm} for the original formulation) for the variational optimization routine. Our method relaxes the assumption in \citet{locatello2017boosting} regarding compact support of variational distributions, allows working with the standard choice of Kullback--Leibler divergence in contrast to \citet{campbell2019universal}, as well as makes assumptions that are strictly milder than Local Asymptotic Normality (LAN) type assumptions in \citet{wang2018frequentist}. Our first result is on understanding statistical properties of the global optimizer of the boosting algorithm. In particular, we show that the Kullback--Leibler divergence of the posterior from the optimal variational solution is bounded in probability, a phenomenon that is similar to what is observed in Bernstein-von Mises theorems for regular parametric models. Our second result pertains to convergence analysis of the algorithm. Our findings are less than encouraging, much along the conjecture of \citet{campbell2019universal}. Specifically, we show that the number of iterations required for the boosting algorithm to converge is exponential in the inverse bandwidth, which is a parameter crucial to the definition of our small bandwidth mixture variational family. We provide intutive justification for this in sections 4.2 and 4.3.

\section{BACKGROUND AND TARGET}
We start with $\mathbb{R}^p$-valued data points $X_1 \dots ,X_n $ which are independently and identically distributed according to density $f(x;\theta)$, where $\theta \in \mathbb{R}^d$, the parameter space. Given a prior density $\pi(\theta)$ on parameter $\theta$, we denote the posterior of $\theta$ as
$$ \pi_n(\theta)=\frac{\prod_{i=1}^n f(X_i ; \theta) \pi(\theta)}{\int \prod_{i=1}^n f(X_i ; \theta) \pi(\theta) d\theta}. $$ Variational boosting works with the following type of distribution family on $\mathbb{R}^d$:
$$\mathcal{Q}=\mathrm{conv}(\Gamma)=\left\{ \sum_{k=1}^K \beta_k \phi_k : \phi_k \in \Gamma, \; \boldsymbol{\beta} \in \Delta^K , \; K\geq 1 \right\}, $$ 
where $\Gamma$ is any family of simple distributions on $\mathbb{R}^d$ and $\Delta^K$ denotes the unit simplex in $\mathbb{R}^K$. The variational family for boosting framework is thus a flexible mixture family. The main aim is to find $q^*_n(\mathcal{Q}) \in \mathcal{Q}$ such that
\begin{equation}
    q^*_n(\mathcal{Q}) = \mathrm{argmin} \{q \in \mathcal{Q} \mid KL(q| |\pi_n)\}, \quad  m_n^*(\mathcal{Q}) = KL(q^*_n(\mathcal{Q})| |\pi_n), 
\end{equation}
where, $KL(a| |b)=\int a(\theta) \log (a(\theta)/b(\theta)) d\theta$ denotes the Kullback--Leibler divergence of density $b$ from  density $a$, both defined on the parameter space. This proceeds through an optimization routine, called the boosting technique. To describe the algorithm, let $\psi^{(k)}_n$ denote the $k$-th iterate in the algorithm for $k \geq 0$. Given $\psi^{(k)}_n$, the next iterate $\psi^{(k+1)}_n$ is obtained by
$$\psi_n^{(k+1)}=(1-\gamma_K) \psi_n^{(k)} + \gamma_k \phi_n^{(k+1)}, $$
where the weight $\gamma_k \in [0,1]$ and $\phi_n^{(k+1)} \in \Gamma$ depend on the boosting approach employed. \citet{locatello2017boosting} proposed the use of Frank--Wolfe algorithm to tackle boosting technique iterates and our setup bears similarity to theirs. Iterates of this kind are also used in \cite{guo2016boosting} and \citet{locatello2018boosting}. Observe that, if $\psi_n^{(k)}$ is already a mixture distribution, every iteration just adds a new component, namely $\phi_n^{(k+1)}$, to the mixture. The Frank--Wolfe algorithm (see appendix; also refer to \citet{jaggi2013revisiting} and \citet{frank1956algorithm}) handles optimization by proceeding exactly in this fashion, and hence is a natural choice as variational algorithm in this case. A quantity $\mathcal{C}_n$ called curvature (see appendix and (15) in section 4.3 for definition), that depends only on the the objective map $q \mapsto KL(q| |\pi_n)$ and domain of optimization $\mathcal{Q}$, plays a crucial role in this algorithm. After initializing with some $\phi_n^{(0)} \in \Gamma$, the $k$-th step of the routine involves finding the new component $\phi^{(k+1)}$ to be added, through a linear minimization routine called linear minimization oracle (LMO). This intermediate step is carried out by solving the LMO approximately in terms of the derivative of the objective map and curvature $\mathcal{C}_n$. Now note that, $q_n^*(\mathcal{Q})$ defined in (1) is a random quantity with respect to data $X_1, \dots X_n$, and so is each iterate of the boosting routine. We aim to provide statistical properties of the random quantities $m_n^*(\mathcal{Q})$ and $\psi^{(k)}_n$ with respect to the true data generating distribution. As a first step,  we show stochastic boundedness of $m_n^*(\mathcal{Q})$ in our theorem 1. We next use theorem 1 from \citet{jaggi2013revisiting} and our result on upper bounding the curvature $\mathcal{C}_n$ (theorem 2) to upper bound the decrements of objective value in the boosting algorithm in terms of the variational family hyper-parameters and sample size $n$. Finally, we tie up the above two results to gain parity of the theoretical minimum and the algorithm. We end with a result (corollary 2) on the order of the required number of boosting updates for a certain degree of error.

\section{STATISTICAL PROPERTIES OF THE VARIATIONAL OPTIMIZER}

\subsection{The small bandwidth mixture Gaussian family}

Recall the definition of $m_n^*(\mathcal{Q})$, the minimum of the objective map $q \mapsto KL(q| |\pi_n)$ over domain $\mathcal{Q}$. Since the function $a \mapsto KL(a| |b)$ has closed sub-level sets, this minimum is attained, i.e. $m_n^*(\mathcal{Q})=KL(q_n^*(\mathcal{Q})| |\pi_n)$ is the minimum corresponding to the domain $\mathcal{Q}$. $m_n^*(\mathcal{Q})$ is a random quantity with respect to data $X_1 \dots X_n$, so we can make probability statements about it with respect to the true data generating distribution. Before we state our stochastic boundedness theorem (section 3.3), we discuss our setup and introduce our variational family. Consider the following restricted Gaussian family for a fixed $c_0$ with $1<c_0<2$, and some $M,\sigma_n>0$:
\begin{equation}
    \Gamma_n= \left\{ N \left( \mu, \sigma^2 I_d \right) : \left\| \mu \right\|_2 \leq M, \; 0 < \sigma_n \leq \sigma \leq c_0^{1/2}\sigma_n \right\}. 
\end{equation}
Denoting by $\mathrm{conv}(\Gamma_n)$ the set of all finite affine combinations of members of $\Gamma_n$, we define $\mathcal{Q}_n = \mathrm{conv}(\Gamma_n)$ as the following restricted mixture Gaussian family:
\begin{equation}
    \mathcal{Q}_n=\left\{ \sum_{k=1}^K \beta_k \phi_k : \phi_k \in \Gamma_n, \; \boldsymbol{\beta} \in \Delta^K , \; K\geq 1 \right\}.
\end{equation}
This domain $\mathcal{Q}_n$ is our variational family of choice, which we call the small bandwidth mixture Gaussian family. Observe that, the components of the mixtures are isotropic Gaussians with means lying in the radius-$M$ compact Euclidean ball around zero in $\mathbb{R}^d$, while the variance $\sigma$ is constrained to be of the same order as $\sigma_n$, the bandwidth parameter. The specific constraint on the constant $c_0$ plays a crucial role in our analysis and will be justified when we study the employed algorithm and its convergence in detail in section 4.

\subsection{Comparison of divergences and Bernstein-von Mises Phenomenon}
The Bernstein-von Mises (BvM) theorem is a well-known frequentist phenomenon for Bayesian posteriors; refer to \citet{van2000asymptotic} for an overview of the BvM phenomenon. It encompasses results about the asymptotic normality of appropriately scaled posterior distributions under regularity conditions on the likelihood and the prior. Local Asymptotic Normality (LAN) assumptions on the likelihood is a pathway to achieving BvM, and \citet{wang2018frequentist} employ it in the context of variational inference. However, BvM results use total variation distance ($d_{TV}$) as the metric, and we wish to focus on Kullback--Leibler discrepancy. In this context, we give a brief comparison of Kullback--Leibler divergence ($KL$) with total variation distance and Hellinger distance ($d_H$). We choose to work with a simplified setting in order to help emphasize our point. Suppose we use a single normal distribution to approximate the posterior. Say  $X_1, \dots ,X_n$ are $d$-vectors, which are independently and identically distributed as $N\left( \theta, \Sigma \right)$, with  $\theta \sim N \left(\mu_0, \Sigma_0 \right) $ and $\mu_0, \Sigma, \Sigma_0$ known. Let $\overline{X}_n$ denote the sample mean, and $\mu_n,\Sigma_n$ the posterior normal's mean and variance respectively. Let $\theta_0$ denote the true parameter. Then the following statements follow from straightforward calculations for Gaussians,

\paragraph{Results:}

\begin{equation}
    KL \left(N\left( \theta_0, n^{-1}\Sigma \right)| |N\left( \mu_n, \Sigma_n \right) \right) \rightsquigarrow \frac{1}{2} \chi^2_d,
\end{equation}
\begin{equation}
    d_{TV}\left(N\left( \mu_n, \Sigma_n \right), N\left( \theta_0, n^{-1}\Sigma \right) \right) \rightarrow 0 \quad \mathrm{a.s.},
\end{equation}
\begin{equation}
    d_{H}\left(N\left( \mu_n, \Sigma_n \right), N\left( \theta_0, n^{-1}\Sigma \right) \right) \rightarrow 0 \quad \mathrm{a.s.},
\end{equation}
\begin{equation}
    KL \left(N\left( \overline{X}_n, n^{-1}\Sigma \right)| |N\left( \mu_n, \Sigma_n \right) \right) \rightarrow 0 \quad \mathrm{a.s.},
\end{equation}

where '$\rightsquigarrow$' denotes weak convergence and a.s stands for almost sure validity with respect to the data generating distribution. Refer to the appendix for proofs of these statements. Since $d_{TV}$ and $d_H$ are equivalent distances, (5) and (6) imply each other, so we just compare (4) with (5). The result in (5) says the posterior comes close to a single Gaussian distribution in total variation, a.s with respect to the data. However the very same distributions are not close in Kullback--Leibler divergence, even in the simplest Gaussian case, as pointed out by (4). This suggests that under Kullback--Leibler divergence, which is a stronger measure of discrepancy, the divergence between the posterior and a deterministic approximator of it should not go down to zero.  We shall see this property in play in our theorem 1.

Now note the comparison of (4) and (7), where, just by changing the centering from the truth (a deterministic quantity) to the sample mean (a random data-dependent quantity), we achieve convergence to zero under Kullback--Leibler divergence. However, this phenomenon is very special to this case, as the correct centering may be computationally impossible to find for complicated posteriors. Hence, (4) is of more practical importance to us than (7) as a statistical statement. In the next section, we present a result similar in flavor to (4), but milder and applicable much more generally.

In general, for the BvM phenomenon to hold, strong regularity conditions are required, which guarantee posterior shape (Gaussian) with high probability with respect to data. However, we wish to include those posteriors in our analysis as well whose shapes are non-Gaussian, making our analysis more general. Works in \citet{kruijer2010adaptive} and \citet{shen2013adaptive} establish approximations of deterministic densities, suitably smooth and exponentially tailed, using Gaussian mixtures. However, if we wish to apply such results to the posterior, which is a random density, we need high probability statements about the smoothness and tail of the posterior, which might tantamount to using hypotheses that the BvM phenomenon demands.

\subsection{Stochastic Boundedness of the Kullback--Leibler Discrepancy}

We now state a theorem about the theoretical minimum Kullback--Leibler divergence $m^*_n(\mathcal{Q}_n)$; see section 2 for definition of $m_n^*$ and section 3.1 for definition of $\mathcal{Q}_n$. Recall that $X_1, \dots X_n$ are independently and identically distributed data points following density $x \mapsto f(x;\theta)$ and $\pi(\theta)$ is the prior on the parameter $\theta \in \mathbb{R}^d$. With $\theta_0 \in \mathbb{R}^d$ denoting the true parameter value, define the log-likelihood ratio $\ell_i(\theta,\theta_0)=\log \left(f(X_i;\theta)/f(X_i;\theta_0)\right)$ for $i=1,\dots ,n$, $L_n(\theta,\theta_0)=\sum_{i=1}^n \ell_i(\theta,\theta_0)$, $KL(\theta_0| |\theta)=-E(\ell_1(\theta,\theta_0))$, $\mu_2(\theta_0| |\theta)=E(\ell_1(\theta,\theta_0))^2$ and $U(\theta)=- \log(\pi(\theta))$ Also, denote by $KL^{(j)}(\theta_0| |\theta)$ and $\mu_2^{(j)}(\theta_0| |\theta)$ for $j=1,2$ the respective derivatives of the maps with respect to the second argument. Let $s_{\max}(A)$ stand for the highest singular value of square matrix $A$. $\| . \|_2$ stands for the $l_2$ norm on $\mathbb{R}^d$. Denote by $\lesssim, \gtrsim$ the corresponding inequalities up to absolute constants. The following assumptions will be required for the theorem: 

\textit{Assumption 1:} The truth $\theta_0$ satisfies $\|\theta_0\|_2 \leq M$.

\textit{Assumption 2:} The variance bound $\sigma_n$ satisfies $\sigma_n \leq n^{-1/2} \leq c_0^{1/2}\sigma_n$ for all $n \geq 1$.

\textit{Assumption 3:} The quantities $KL(\theta_0| |\theta), \mu_2(\theta_0| |\theta)$ are finite for every $\theta \in \mathbb{R}^d$.

\textit{Assumption 4:} Matrices $KL^{(2)}(\theta_0| |\theta), \mu_2^{(2)}(\theta_0| |\theta)$ and $U_2^{(2)}(\theta)$ exist on $\mathbb{R}^d$ and satisfy for any $\theta, \theta' \in \mathbb{R}^d$:
    \begin{equation}
    \begin{split}
        & s_{max} \left( KL^{(2)}(\theta_0| |\theta) - KL^{(2)}(\theta_0| |\theta') \right) \lesssim \left\| \theta - \theta' \right\|_2^{\alpha_1}, \\
        & s_{max} \left( \mu_2^{(2)}(\theta_0| |\theta) - \mu_2^{(2)}(\theta_0| |\theta') \right) \lesssim \left\| \theta - \theta' \right\|_2^{\alpha_2}, \\
        & s_{max} \left( U_2^{(2)}(\theta) - U_2^{(2)}(\theta') \right)\lesssim \left\| \theta - \theta' \right\|_2^{\alpha_3}, 
        \end{split}
    \end{equation}
    for some $\alpha_1,\alpha_2,\alpha_3 \geq 0$.

\textit{Assumption 5:}
$KL(\theta_0| |\theta) \gtrsim \left\| \theta - \theta_0 \right\|_2^2. $

\textbf{Theorem 1:} \textit{Under assumptions 1-5, it holds that}   $m^*_n(\mathcal{Q}_n)$ \textit{is bounded in probability with respect to the data generating distribution, i.e. given any} $\epsilon \in (0, 1)$, \textit{there exists} $M_{\epsilon},N_{\epsilon} > 0$ \textit{such that for all} $n\geq N_{\epsilon}$, we have with probability greater than $1 - \epsilon$
\begin{equation}
    m^*_n(\mathcal{Q}_n) < M_{\epsilon}. 
\end{equation}

\textit{Remark 1:} Assumption 2 dictates the exact order of $\sigma_n$, and also allows $N_\epsilon=1$. It is at par with the order of the bandwidth expected in parametric estimation. Finiteness of $\mu_2(\theta_0| |\theta)$ in assumption 3 is crucial for concentration inequalities to be applied. The smoothness assumption i.e. assumption 4, helps dictate posterior moments, but not the shape of the entire posterior. The final assumption is the standard identifiability condition for using Kullback--Leibler divergence. Assumptions of these types are quite common in the literature; refer to the moment assumptions for the posterior in Huggins et al. [2020]  in the context of distributional bounds in variational inference and section 5 in Ghosal et al. [2000] in the context of posterior contraction. An important observation is the fact that we do not intend to recover the posterior mean as in \citet{pati2018statistical}, where assumptions are aimed at studying variational point estimates.

\textit{Remark 2:} The hypothesis for the theorem is milder than what can guarantee a weak convergence type result, like the (4) in section 3.2. There are no assumptions that allow local quadratic nature of the posterior, as we impose conditions only on the expected log-likelihood. This makes our assumptions more general than a BvM type setup. Contrast this with the stochastic LAN assumption in \citet{wang2018frequentist}, which approximates the likelihood with a stochastic linear term and a deterministic quadratic term.

\textit{Remark 3:} Theorem 1 establishes $m^*_n(\mathcal{Q}_n)$ to be an $O_p(1)$ quantity with respect to the true data generating distribution. From the proof in appendix, one can further conclude $M_{\epsilon} \gtrsim \epsilon^{-1/2}$ for small $\epsilon$.

We shall state a corollary which is a simplification of theorem 1 in the case the density $x \mapsto f(x:\theta)$ belongs to the $K$-parameter exponential family on $\mathbb{R}^p$. Let $\theta \in \mathbb{R}^d$ be the canonical parameter, $T_l : \mathbb{R}^p \mapsto \mathbb{R}, \; l=1, \dots K$ be the sufficient statistics and $A:\mathbb{R}^d \mapsto \mathbb{R}$ be the log-partition function. The form of the density is given by
$$f(x;\theta)=\exp \left( \sum_{l=1}^K \theta_l T_l(x) -A(\theta) \right). $$
Let $A^{(j)}(\theta), j=1,2$ denote the respective derivatives of $A(\theta)$. We shall use the notion of strong convexity in the corollary that follows; refer to the appendix for a general definition. We also need to note down the definition of $\alpha$-Lipschitz functions:

\textit{Definition:} Vector or square-matrix valued functions $f$ defined on $D \subset \mathbb{R}^d$ are said to be $\alpha$-Lipschitz for an $\alpha > 0$, if there exists a constant $C>0$ such that for all $x, y \in D$
\begin{enumerate}
    \item $\|f(x)-f(y)\|_2 \leq C \|x-y\|^{\alpha}_2$ for vector valued functions, and
    \item $s_{\max}(f(x)-f(y)) \leq C \|x-y\|^{\alpha}_2$ for square-matrix valued functions.
\end{enumerate}

\textbf{Corollary 1:} \textit{Assume that} $A(\theta)$ \textit{is twice differentiable and strongly convex on} $\mathbb{R}^d$. \textit{Also assume that the vectors} $A^{(1)}(\theta), A^{(2)}(\theta) \theta$ \textit{and the square matrices} $A^{(1)}(\theta) A^{(1)}(\theta)^T, A^{(2)}(\theta)$ \textit{are} $\alpha$-\textit{Lipschitz functions of} $\theta \in \mathbb{R}^d$ \textit{for some} $\alpha \geq 0$. \textit{Under these conditions, assumptions on the expected likelihood (assumption 3 and 4) in theorem 1 hold.}

\subsection{Sketch of Proof of Theorem 1}
We now briefly discuss the proof technique of theorem 1, shedding more light on the importance of the assumptions made; refer to appendix for a detailed proof of theorem 1. Note that $m^*_n(\mathcal{Q}_n)$ is bounded above by the objective map $q \mapsto KL(q| |\pi_n)$ evaluated at any member of $\mathcal{Q}_n$. We choose that member to be $q_0$, the $d$-dimensional Gaussian density centered at the truth $\theta_0$ and variance $\sigma_n^2 I_d$. Here, $I_d$ stands for the $d$-dimensional identity matrix and  $\sigma_n$ satisfies assumption 2. Along with assumption 1, we have, $q_0 \in \mathcal{Q}_n$ and hence $m^*_n(\mathcal{Q}_n) \leq KL(q_0| |\pi_n)$. Thus it is enough to show $KL(q_0| |\pi_n)$ is bounded in probability. This Kullback--Leibler discrepancy can now be broken down in a sum to give two deterministic and two random quantities. The stochastic part of the sum is given by
$$\log \left( m(X_n) \right) - \left( \int L_n(\theta,\theta_0) q_0(\theta) d\theta \right), $$
where
$$m(X_n) = \int \exp \left( L_n(\theta,\theta_0) \right) \pi(\theta) d\theta. $$
Under true data generating distribution, the integrand defining $m(X_n)$ has expectation $1$, which helps upper bound in probability the first stochastic term above. For the second, we notice that $L_n(\theta,\theta_0)$ has expectation $-n KL(\theta_0| |\theta)$ under true $\theta_0$. We then utilize the smoothness and identifiability assumptions on $KL(\theta_0| |\theta)$ to obtain its Taylor series expansion around $\theta_0$, that helps lower bound in probability the second stochastic term through Chebyshev's inequality. We can now conclude our result by noting that sum of $O_p(1)$ quantities is again $O_p(1)$.

\section{CONVERGENCE ANALYSIS OF THE ALGORITHM}

\subsection{Steps of the Algorithm}

We opt for a functional version of the Frank--Wolfe algorithm as our variational boosting algorithm (algorithm 1), which bears analogy to variant 0 of algorithm 1 in \citet{locatello2017boosting}. Refer to \citet{jaggi2013revisiting} for the general algorithm and our discussion in the appendix for a brief overview and notations. Recall that we aim to perform the optimization (1) with domain $\mathcal{Q}=\mathcal{Q}_n=\mathrm{conv}(\Gamma_n)$. Thus our objective map is $q \mapsto KL(q| |\pi_n), q \in \mathcal{Q}_n$, which is convex on its domain. Let us generically denote members of $\Gamma_n$, which are single Gaussians, by $\phi$ and those of $\mathcal{Q}_n$, which are mixture Gaussians, by $\psi$. Superscripts stand for iterate numbers and the subscript $n$ denotes dependence on sample size. Say $\psi_n^{(k)}$ is the mixture obtained as the $k$-th step iterate. We use the notation $\mathcal{D}_n$ to denote the Bregman divergence of our convex objective map (refer to appendix for general definition of Bregman divergence). For $\psi_1, \psi_2 \in \mathcal{Q}_n$, the Bregman divergence $\mathcal{D}_n$ of $\psi_1$ from $\psi_2$, at $\psi_1$, under the objective map, is given by
\begin{equation}
    \mathcal{D}_n(\psi_2| |\psi_1)=KL(\psi_2| |\pi_n)-KL(\psi_1| |\pi_n) - \int \left( \psi_2 - \psi_1 \right) \left( \log (\psi_1) - \log (\pi_n) \right) d \theta. 
\end{equation}
The last term above derives from the fact that our domain of optimization, $\mathcal{Q}_n$, lies embedded in the $L_2$ inner product function space on $\mathbb{R}^d$. The term $\log (\psi_1) - \log (\pi_n)$, appearing within the integrand in (10), is the sub-gradient (see appendix for general definition of subgradient) of the objective map at $\psi_1$, and $\psi_n^{(k)}$ plays the role of $\psi_1$ at the $k$-th step of the algorithm. We use the notation $\mathcal{C}_n$ to denote the curvature of the objective map for the domain $\mathcal{Q}_n$ (see section 4.3 for details on $\mathcal{C}_n$). 

\begin{algorithm}[tb]
\caption{Functional Frank--Wolfe Algorithm with small bandwidth mixture Gaussian variational family}
\label{algorithm1}
    \begin{enumerate}
        \item Initialize with $\psi_n^{(0)}=\phi_n^{(0)} \in \Gamma_n$.
        \item At $k$-th step, calculate the subgradient $s_n^{(k)}=\log (\psi_n^{(k)}) - \log (\pi_n)$ at the $k$-th iterate $\psi_n^{(k)}$.
        \item Set $\gamma_k=2/(k+2)$, solve LMO approximately i.e. find $\phi_n^{(k+1)} \in \Gamma_n$ such that $\int s_n(\theta) \phi_n^{(k+1)} d\theta \leq \min \left\{\phi \in \Gamma_n \mid \int s_n^{(k)} \phi(\theta) d\theta \right\} + \gamma_k \mathcal{C}_n / 2$.
        \item Update $\psi_n^{(k+1)}=(1-\gamma_k)\psi_n^{(k)}+\gamma_k \phi_n^{(k+1)}$ to get the $(k+1)$-th iterate.
    \end{enumerate}
\end{algorithm}

That $s_n^{(k)}=\log (\psi_n^{(k)})-\log (\pi_n)$ in algorithm 1 is indeed a valid subgradient of the objective map at $\psi_n^{(k)}$, follows from the following lemma on Bregman divergence $\mathcal{D}_n$ and the non-negativity of Kullback--Leibler divergence. It basically says calculating $\mathcal{D}_n$ and Kullback--Leibler divergence of the objective map are one and the same.

\textbf{Lemma 1:} \textit{For any densities} $\psi_1$ \textit{and} $\psi_2$ \textit{defined on} $\mathbb{R}^d$, \textit{we have} $$\mathcal{D}_n(\psi_2| |\psi_1)=KL(\psi_2| |\psi_1). $$

In algorithm 1, the target is to greedily fit single Gaussian components to build a mixture of Gaussians, that is close to the posterior in Kullback--Leibler divergence. Let the optimal approximating mixture be denoted by $\psi^{*}_n = q_n^{*}(\mathcal{Q}_n)$ (see (1) for definition of $q_n^{*}$). Note that, both $\psi_n^{(k)}$ and $\psi^{*}_n$ are random with respect to data. But the practitioner is given the data $X_1, \dots X_n$, hence she runs algorithm 1 deterministically and upper bounds, point-wise on the sample space, the random quantity 
$$KL(\psi_n^{(k)}||\pi_n)-KL(\psi_n^{*}||\pi_n). $$ 
We want to find the aforementioned upper bound in terms of number of iterations $k$ and sample size $n$.

Step 3 of algorithm 1 is an approximate linear minimization routine for which we shall use the parametric structure of Gaussians and optimize over the parameters. The practitioner starts with an initial guess of $\mu^{(0)},  \sigma^{(0)}$ such that $N\left( \mu^{(0)}, (\sigma^{(0)})^2 I_d \right) \in \Gamma_n$. This gives her the Gaussian $\phi_n^{(0)}=N\left( \mu^{(0)}, (\sigma^{(0)})^2 I_d \right)$. At the beginning of the $k$-th step, she has the mixture $\psi_n^{(k-1)}=\sum_{j=1}^{k-1} \beta_j \phi_n^{(j)} $, where the vector $\beta \in \Delta^{k-1}$ and are positive functions of $\gamma_l$'s with $\gamma_l=2/(l+2), l=1 \dots k-1$. In order to obtain the Gaussian $\phi_n^{(k)}$, she finds  approximate $\mu^{(k)}, \sigma^{(k)}$ through the LMO optimization routine
\begin{equation}
    \mathrm{argmin} \Bigg\{ \left\| \mu \right\|_2 \leq M, \; 0 < \sigma_n \leq \sigma \leq \sqrt{c_0}\sigma_n : \int \phi(\theta; \mu, \sigma^2) \log \left( \frac{\psi_n^{(k-1)}(\theta)}{\pi_n(\theta)} \right) d\theta \Bigg\}.
\end{equation}

By lemma 5 of \citet{jaggi2013revisiting}, the practitioner is allowed to use any algorithm at her disposal for the above routine, as long as she is able to perform the optimization of this $k$-th step with error $\leq \gamma_k \mathcal{C}_n/2$. A very important observation is  that knowing the normalizing constant of the posterior is not necessary for the above routine. Thus, using the $\phi_n^{(k)}$ obtained, we update $\psi_n^{(k)}=(1-\gamma_k)\psi_n^{(k-1)} + \gamma_k \phi_n^{(k)}$. 

\subsection{Rate of Convergence of Algorithm 1}

We now state our main theorem on rate of convergence:

\textbf{Theorem 2:} 
\begin{equation}
    KL(\psi_n^{(k)} ||\pi_n) - KL(\psi^{*}_n||\pi_n) \leq \frac{8 (2-c_0)^{-d/2} \exp \left( \frac{2 M^2}{(2-c_0)\sigma_n} \right)}{k+2}.
\end{equation}
This theorem upper bounds the gap in value of the Kullback--Leibler objective map, between the $k$-th boosting iterate and the optimal approximator to the posterior. It depends upon the sample size, number of iterations,  dimension of parameter space and hyper-parameters of the domain $\mathcal{Q}_n$. We make the following important remarks:

\textit{Remark 4:} The above convergence-rate holds point-wise with respect to the data generating distribution, and the upper bound is deterministic.

\textit{Remark 5:} The rate is sub-linear in the number of iterates $k$ and exponential in inverse bandwidth $\sigma_n^{-1}$. Sub-linearity follows from the use of Frank--Wolfe, while exponentiality is an artifact of enforcing Kullback--Leibler divergence to be strongly smooth over our small bandwidth domain.

We now combine theorems 1 and 2 to obtain a novel probability statement about the random $k$-th iterate of our boosting algorithm. It is important to know how many iterations we need in order to obtain a certain error in the algorithm, and that depends on the sample size $n$. So we let $k$ vary with sample size $n$, i.e. take $k \equiv k_n$.

\textbf{Corollary 2:} \textit{If} $k_n \gtrsim \exp(n^{1/2})$, \textit{then for any} $\epsilon \in (0, 1)$, \textit{there exists constant} $C>0$, \textit{such that with probability greater than $1 - \epsilon$ we have}
\begin{equation}
KL(\psi_n^{(k_n)}||\pi_n) < C \epsilon^{-1/2}.
\end{equation}

\textit{Remark 6:}
Corollary 2 shows the the required order of $k_n$, the number of iterations, in terms of sample size $n$ that maintains order parity while combining theorems 1 and 2. It shows after how many runs of the boosting algorithm the random iterates are guaranteed to be bounded in probability with respect to the data generating distribution.

\subsection{Sketch of Proof of Theorem 2}

We briefly discuss how we arrived at theorem 2. For $k$-th step, theorem 1 of \citet{jaggi2013revisiting} allows the following upper bound regarding our objective map:
\begin{equation}
    KL(\psi_n^{(k)}||\pi_n)-KL(\psi^{*}_n||\pi_n) \leq \frac{4 \mathcal{C}_n}{k+2}, 
\end{equation}
where $\mathcal{C}_n$ denotes the curvature of the objective map $q \mapsto KL(q| |\pi_n)$ over domain of optimization $\mathcal{Q}_n$. It is given by

\begin{equation}
    \mathcal{C}_n = \sup \Bigg\{ \frac{2}{\alpha^2} \mathcal{D}_n (\psi_2| |\psi_1) : \psi_1 \in \mathcal{Q}_n, \phi \in \Gamma_n, \alpha \in [0,1] , \psi_2 = \psi_1 + \alpha (\phi - \psi_1) \Bigg\}.
\end{equation}
Curvature is thus essentially the maximum scaled Bregman divergence between densities in domain $\mathcal{Q}_n$ and their perturbations through mixtures. It is entirely determined by $n, \sigma_n$ and the variational family hyper-parameters $M, c_0, d$. Refer to the appendix for a general definition of curvature. We prove the following lemma that upper bounds $\mathcal{C}_n$ on the small bandwidth mixture Gaussian domain $\mathcal{Q}_n$:

\textbf{Lemma 2:}
\begin{equation}
    \mathcal{C}_n \leq 2 (2-c_0)^{-d/2} \exp \left( \frac{2 M^2}{(2-c_0)\sigma_n} \right).
\end{equation}    

\textit{Remark 6:} Note the exponential dependence of the bound on inverse bandwidth $\sigma_n^{-1}$. Whether $\mathcal{C}_n$ is a finite quantity depends on the constraints in the definition of $\Gamma_n$ (see section 3.1) through which it is defined. We are enforcing finiteness of $\mathcal{C}_n$ through utilizing proposed small bandwidth mixture Gaussian family. which is the essence of the constraint $1 < c_0 < 2$ in the definition of $\Gamma_n$.

\textit{Remark 7:} \citet{locatello2017boosting} and our work share the central motive of ensuring strong smoothness (see definition in appendix) of Kullback--Leibler divergence, which essentially dictates finiteness of $\mathcal{C}_n$. It is a property of both the objective function to be minimized, and the domain over which it is minimized. However,  \citet{locatello2017boosting} assume $\Gamma_n$ to consist of truncated, lower bounded densities, which although works from a practical point of view, excludes the very basic Gaussian distribution. Their theory only accommodates compactly supported densities, which is too restrictive and is remedied through the above lemma. Thus lemma 2 is a cardinal contribution of this work; it shows that the curvature of Kullback--Leibler discrepancy is bounded over the domain of small bandwidth Gaussian mixture family.

\textit{Remark 8:} If $c_0 \geq 2$, then upper bound on $\mathcal{C}_n$ improves with smaller bandwidth $\sigma_n$, which is evidently untrue, as smaller bandwidth Gaussian mixtures are spikier and worse approximators through Kullback--Leibler divergence. If $c_0 < 1$, the bound improves for higher dimension, which contradicts curse of dimensionality. This intuitively justifies the technical importance of the constraint: $1 < c_0 < 2$.

We now state a straightforward formula that shall play a crucial role in the proof of lemma 2, as well as provide insight into why the proposed family of small bandwidth Gaussian mixtures is a good choice as a variational family.

\textbf{Lemma 3:} \textit{Whenever} $2\sigma_1^2 > \sigma_2^2 > 0$, $\chi^2$ \textit{distance between two Gaussians is given by}
\begin{equation}
    \chi^2 \left(\mathrm{N} \left(\mu_2,\sigma_2^2 \mathbf{I}_d\right)|| \; \mathrm{N} \left(\mu_1,\sigma_1^2 \mathbf{I}_d\right)\right) = -1 + \left( \frac{\sigma_1^2}{\sigma_2 \sqrt{2\sigma_1^2 - \sigma_2^2}} \right)^d \exp \left( \frac{\| \mu_2 - \mu_1 \|_2^2}{2\sigma_1^2 - \sigma_2^2} \right).
\end{equation}

\textit{Remark 8:} The $\chi^2$ distance (refer to appendix for definition) arises during the calculations of curvature of Kullback--Leibler divergence, and one can clearly see the right hand side above is defined as a real number only for a certain interval of values of the variances. This calculation was an important motivator for the small bandwidth variational family. It is also the reason for the exponential dependence of the upper bound, in theorem 2, on the inverse bandwidth.

We can now directly plug into lemma 3 the domain parameters of $\mathcal{Q}_n$ to calculate the proposed upper bound in lemma 2. In turn, lemma 2 directly gives us theorem 2 in combination with (14).

\section{CONCLUDING REMARKS}
To the best of our knowledge, we provide, for the first time, statistical properties of the iterates in a variational boosting algorithm. As part of frequentist validation of this variational method, we assume regularity conditions similar to ones widely used in Bayesian contraction literature. Our hypotheses are general enough to include most likelihoods and priors; they do not demand Gaussian like posteriors. Regarding the boosting algorithm itself, we employ the non-compact domain of mixture Gaussian densities as our variational family, much in contrast to the use of compact domain in \citet{locatello2017boosting}. Convergence analysis of our algorithm shows how smoothness properties of Kullback--Leibler discrepancy lead to exponential dependence of iterate number on the inverse bandwidth, which can be contrasted to faster convergence rates for weaker metrics like Hellinger distance \citep{campbell2019universal}. 

\section{SUPPLEMENTARY MATERIALS (APPENDIX)}
\section*{REVISITING FRANK--WOLFE ALGORITHM}
\begin{algorithm}
\caption{Frank--Wolfe algorithm with approximate Linear Minimization Oracle:}
    \begin{enumerate}
    \item Initialize with $x^{(0)} \in D$.
    \item For the $k$-th step, set $\gamma_k=2/(k+2)$ and calculate sub-gradient $\nabla f(x^{(k)})$.
    \item Solve the linear minimization oracle (LMO) approximately, i.e. find $y^{(k+1)} \in D$ such that $\langle \nabla f(x^{(k)}), y^{(k+1)} \rangle \leq \min \left\{y \in D \mid \langle \nabla f(x^{(k)}), y \rangle \right\} + \gamma_k \mathcal{C}_{f,D} / 2$.
    \item Update $x^{(k+1)}=(1-\gamma_k)x^{(k)}+\gamma_k y^{(k+1)}$.
\end{enumerate}
\end{algorithm}

We note down the basics of Frank--Wolfe algorithm in this section. The reader is referred to \cite{jaggi2013revisiting} for further details and to \cite{frank1956algorithm} for the original formulation. We start by reviewing the notation. In what follows, $Y$ is an inner product space with $\langle y_1,y_2 \rangle$ denoting the inner product of $y_1,y_2 \in Y$ and $\|y\|=\langle y, y \rangle$ the norm induced by the inner product. $D$ shall denote a compact, convex subset of $Y$, which is our optimization domain. We shall work with a convex function $f$ defined on $D$, which is our objective function for the optimization routine. We start by revising the notion of a subgradient. \newline

\textbf{Definition 1:} A sub-gradient of $f$ at $x \in D$, denoted by $\nabla f(x)$, is a member of $\partial f(x) \subset D$, given by
$$\partial f(x)= \left\{ y \in D \mid f(z)-f(x)-\langle y, z-x \rangle \geq 0, \; \forall \, z \in D \right\}.$$
It is easy to note that, if $Y=\mathbb{R}^d$, and convex function $f$ is differentiable at $x \in Y$, then the gradient at $x$ satisfies $f'(x) \in \partial f(x)$. Subgradients are useful when the notion of differentiability is untenable. Next, we note down the definition of Bregman divergence. \newline

\textbf{Definition 2:}
For any $x,y \in D$, the Bregman divergence of $y$ from $x$ under function $f$ is defined as
\begin{equation}
    \mathcal{D}_{f}(y| |x)=f(y)-f(x)-\langle \nabla f(x), y-x \rangle.
\end{equation}
With this definition in hand, we now define the curvature of $f$ on domain $D$. \newline

\textbf{Definition 3:} The curvature $\mathcal{C}_{f,D}$ of $f$ on domain $D$ is given by
\begin{equation}
    \mathcal{C}_{f,D}= \sup \left\{ \frac{2}{\alpha^2} \mathcal{D}_f (x_2| |x_1): y, x_1 \in D, \alpha \in [0,1] , x_2 = x_1 + \alpha (y - x_1) \right\}.
\end{equation}
One can think of the curvature as the maximum scaled Bregman divergence between points in $D$ and their perturbations through mixtures. We now recall the definition and significance of strong smoothness and strong convexity of $f$. \newline

\textbf{Definition 4:}
If for any $x ,y \in D$ and some $C_1, C_2 > 0$ (possibly depending on $f$ and $D$)
\begin{enumerate}
    \item $\mathcal{D}_f(y| |x) \leq C_1 \|y-x\|^2$, then $f$ is strongly smooth on $D$,
    \item $\mathcal{D}_f(y| |x) \geq C_2 \|y-x\|^2$, then $f$ is strongly convex on $D$.
\end{enumerate}
Convex functions $f$ on $D$ that satisfy strong smoothness allow calculations of rate of convergence. The most basic Frank--Wolfe algorithm minimizes convex function $f$, defined on domain $D$. We note down a version with approximately solved subproblem, following \cite{jaggi2013revisiting}. \newline

Let $x^*$ denote the minimum point in domain $D$. The above algorithm, by theorem 1 in \cite{jaggi2013revisiting}, satisfies
\begin{equation}
    f(x^{(k)})-f(x^*) \leq 4 C_{f,D}/(k+2)
\end{equation}
This gives us the rate of convergence of this algorithm, in terms of the curvature $\mathcal{C}_{f,D}$. Note that such a rate of convergence with respect of number of iterations $k$ is called sub-linear. For statistical problems, $\mathcal{C}_{f,D}$ is typically a function of the sample size and the parameter dimension, and can be quite large for densities having non-compact support.  \newline

\section*{PROOF OF THEOREM 1}

In what follows, $\gtrsim, \lesssim$ respectively stand for greater than and less than up to an absolute constant and $s_{\max}(A)$ denotes the highest singular value of square matrix $A$. Let $q_0$ denote the $d$-dimensional Gaussian density, centered at the truth $\theta_0$, and variance $\sigma_n^2 I_d$, where $\sigma_n$ satisfies assumption 2. Along with assumption 1, we have $q_0 \in \mathcal{Q}_n$ and hence $m^*_n(\mathcal{Q}_n) \leq KL(q_0| |\pi_n)$, so that it is enough to show $KL(q_0| |\pi_n)$ is bounded in probability. We decompose $KL(q_0| |\pi_n)$ as
\begin{equation}
    \int q_0(\theta) \log \frac{q_0(\theta)}{\pi_n(\theta)}d\theta = -d\left[\log(\sqrt{2\pi}) + \frac{1}{2}\right] + \log \left( m(X_n) \right) - \left( \int L_n(\theta,\theta_0) q_0(\theta) d\theta \right) + \int U(\theta) q_0(\theta) d\theta.
\end{equation}
Since sum of $O_p(1)$ quantities are again $O_p(1)$, we can stochastically bound the above expression term by term.
The first and last terms are already constants. We will also prove $\int U(\theta) q_0(\theta) d\theta \lesssim U(\theta_0) $. So, it is enough to show $\log \left( m(X_n) \right)$ and $\int L_n(\theta,\theta_0) q_0(\theta) d\theta$ are stochastically bounded from above and below, respectively. We have
\begin{equation}
    \mathrm{pr} \left( \log \left( m(X_n) \right) > -\log \epsilon  \right) = \mathrm{pr} \left( \left( m(X_n) \right) > \frac{1}{\epsilon}  \right) \leq \epsilon . E \left( m(X_n) \right) = \epsilon \quad \forall \, n.
\end{equation}
We now employ Taylor expansion around $\theta_0$. Using assumption 4 for $U(\theta)$, observe that
\begin{equation}
    \int q_0(\theta) U(\theta)d\theta \lesssim U(\theta_0) + s^2_{max}\left( U^{(2)}(\theta_0) \right) \left( \int \left\| \theta - \theta_0 \right\|^2 q_0(\theta) d\theta \right) + \left( \int \left\|\theta - \theta_0   \right\|_2^{2+2\alpha_3} q_0(\theta) d\theta \right) \lesssim U(\theta_0),
\end{equation}
and for $\mu_2(\theta_0| |\theta)$, observe that
\begin{equation}
    \int q_0(\theta) \mu_2(\theta_0| |\theta) d\theta \leq s^2_{max} \left( \mu_2^{(2)}(\theta_0| |\theta_0) \right) \left( \int \left\| \theta - \theta_0 \right\|^2 q_0(\theta) d\theta \right) + \left( \int \left\|\theta - \theta_0   \right\|_2^{2+2\alpha_2} q_0(\theta) d\theta \right) \leq C_1 \sigma_n^2,
\end{equation}
for constant $C_1 > 0$.
Again, by assumption 4 for $KL(\theta_0| |\theta)$, we get the upper bound (similar to previous step)
\begin{equation}
    \int q_0(\theta) KL(\theta_0| |\theta)d\theta \leq C_u \sigma_n^2,
\end{equation}
and using assumption 5, we get the lower bound
\begin{equation}
    C_l \sigma_n^2 \leq \int q_0(\theta)  KL(\theta_0| |\theta)d\theta,
\end{equation}
where $0 < C_l < C_u$ are absolute constants.
Now, for $\delta >0$ to be chosen later, we have
\begin{equation}
\begin{split}
    &\hspace{15pt} \mathrm{pr}  \left[ \int L_n(\theta,\theta_0) q_0(\theta) d\theta \leq - C_u (1+\delta) n \sigma_n^2 \right] \\
    &\leq \mathrm{pr}  \left[ \int L_n(\theta,\theta_0) q_0(\theta) d\theta \leq - (1+\delta) n \int KL(\theta_0| |\theta) q_0(\theta) d\theta \right] \\
    &\leq \mathrm{pr}  \left[ \int \frac{1}{\sqrt{n}} \zeta_n(\theta,\theta_0) q_0(\theta) d\theta \leq - \delta \sqrt{n} \int KL(\theta_0| |\theta) q_0(\theta) d\theta\right] \\
    &\leq \frac{E\left( \int l(\theta,\theta_0) q_0(\theta) d\theta \right)^2}{\delta^2 n \left( \int KL(\theta_0| |\theta) q_0(\theta) d\theta\right)^2} \leq \frac{\int q_0(\theta) \mu_2(\theta_0| |\theta) d\theta}{\delta^2 n \left( \int KL(\theta_0| |\theta) q_0(\theta) d\theta\right)^2}\leq \frac{C_1 \sigma_n^2}{\delta^2 C_l^2 n \sigma_n^4}=\frac{C_1}{\delta^2 C_l n\sigma_n^2}.
\end{split}
\end{equation}
From assumption 2, we have $c_0^{-1/2} n^{-1/2} \leq \sigma_n \leq n^{-1/2}$ and hence, given $\epsilon > 0$, we can choose $\delta:=\left(C_1 c_0^{1/2} / \epsilon C_l\right)^{1/2}$ to get
\begin{equation}
        \mathrm{pr}  \left[ \int L_n(\theta,\theta_0) q_0(\theta) d\theta \leq-C_u \left(1+\left(C_1 c_0^{1/2} / \epsilon C_l\right)^{1/2}\right) \right] \leq \epsilon, 
\end{equation}
and the stochastic boundedness result is complete.

\section*{PROOF OF COROLLARY 1}

For this specific family of densities, let $\nu_l = E \left( T_l |\theta_0 \right), \; l=1. \dots K$ denote the expectations of the sufficient statistics, and let $\sigma_{\ell_1, l_2}$ denote $Cov(T_{\ell_1}, T_{l_2} | \theta_0)$ for $\ell_1, l_2 = 1, \dots K$. Let $\nu_0 = (\nu_1, \dots \nu_K)$ and $\Sigma_0 = ((\sigma_{\ell_1, l_2}))_{\ell_1, l_2}$, so that for the $K$-vector $T = (T_1, \dots T_K)$, $E(T|\theta_0)=\nu_0$ and $Cov(T|\theta_0) = \Sigma_0$. Direct calculations shows
\begin{equation}
    KL(\theta_0| |\theta) = A(\theta)-A(\theta_0)-(\theta - \theta_0)^T \nu_0, 
\end{equation}
and we know that for Exponential families, $\nu_0 = A^{(1)}(\theta_0)$. Thus (12) becomes (see Definition 2)
\begin{equation}
    KL(\theta_0| |\theta) = D_A(\theta_0| |\theta).
\end{equation}
It is worthwhile to observe the analogy to lemma 2. $KL(\theta_0| |\theta)$ is finite for all $\theta \in \Theta$ as $D_A(\theta_0| |\theta)$ is. Now by hypothesis, $A(\theta)$ is strongly convex, and hence by part 2 of Definition 4 and (13), we have $KL(\theta_0| |\theta) \geq C \| \theta - \theta_0 \|_2^2$ for some constant $C > 0$. Next, calculation shows
\begin{equation}
    \mu_2(\theta_0| |\theta) = (\theta - \theta_0)^T \Sigma_0 (\theta - \theta_0) + \left( D_A(\theta_0| |\theta) \right)^2,
\end{equation}
and we know that for Exponential families, $\Sigma_0 = A^{(2)}(\theta_0)$. Thus (14) becomes
\begin{equation}
    \mu_2(\theta_0| |\theta) = (\theta - \theta_0)^T A^{(2)}(\theta_0) (\theta - \theta_0) + \left( D_A(\theta_0| |\theta) \right)^2.
\end{equation}
Since, by hypothesis, $A^{(2)}(\theta)$ exists and $D_A(\theta)$ is finite for all $\theta \in \Theta$, we conclude $\mu_2(\theta_0| |\theta)$ is finite for all $\theta \in \Theta$. This verifies assumptions 3 and 5 starting from the hypothesis. Using (13), we see that the first and second derivatives of $D_A$,  with respect to the second argument, satisfy
\begin{equation}
    D_A^{(1)}(\theta_0| |\theta) = A^{(1)}(\theta) - A^{(1)}(\theta_0), \quad D_A^{(2)}(\theta_0| |\theta) = A^{(2)}(\theta).
\end{equation}
Thus, from (30), (32) and (33), we have
\begin{equation}
\begin{split}
    KL^{(2)}(\theta_0| |\theta) &= A^{(2)}(\theta), \\
    \mu_2^{(2)}(\theta_0| |\theta) = 2 \left( A^{(2)}(\theta_0) + D_A(\theta_0| |\theta) A^{(2)}(\theta) \right) &+ \left( A^{(1)}(\theta) - A^{(1)}(\theta_0) \right) \left( A^{(1)}(\theta) - A^{(1)}(\theta_0) \right)^T.
\end{split}
\end{equation}
Since sum of $\alpha$-Lipschitz functions is again $\alpha$-Lipschitz, (17) shows why the hypothesis of the corollary suffices to conclude that assumption 4 of theorem 1 holds. 

\section*{PROOF OF LEMMA 1}

 $KL(\psi_2||\pi_n) = \int \psi_2 \log \psi_2 - \int \psi_2 \log \pi_n$, $KL(\psi_1||\pi_n) = \int \psi_1 \log \psi_1 - \int \psi_1 \log \pi_n$, so that $KL(\psi_2||\pi_n) - KL(\psi_1||\pi_n)=KL(\psi_2||\psi_1) + \int (\psi_2-\psi_1) \log \psi_1 - \int (\psi_2 - \psi_1) \log \pi_n = KL(\psi_2||\psi_1) + \int (\psi_2-\psi_1) (\log \psi_1 - \log \pi_n)$ and we are done.

\section*{PROOF OF LEMMA 2}

Let $\phi, \phi_1, \dots \phi_k \in \Gamma_n, \psi_1 = \sum_{j=1}^n \beta_j \phi_j$ for $\boldsymbol{\beta} \in \Delta^k$ and $\psi_2 = \psi_1 + \alpha(\phi - \psi_1)$ for $\alpha \in [0,1]$. Starting with lemma 1, we have the Taylor expansion
\begin{equation}
    \begin{split}
        \mathcal{D}_n(\psi_2| |\psi_1) &= KL(\psi_1 + \alpha(\phi - \psi_1)||\psi_1) \\
        & = \alpha \frac{\partial}{\partial \alpha'} \bigg|_{\alpha'=0} KL(\psi_1 + \alpha'(\phi - \psi_1)||\psi_1) + \frac{\alpha^2}{2} \frac{\partial^2}{\partial \alpha^{'2}} \bigg|_{\alpha'=\beta} KL(\psi_1 + \alpha'(\phi - \psi_1)||\psi_1) \\
        & =: \alpha T_1 + \frac{\alpha^2}{2}T_2,
    \end{split}
\end{equation}
where $\beta$ lies in between $0$ and $\alpha$, and $T_1, T_2$ denote the first and second derivatives respectively. For $T_1$ we have
\begin{equation}
    \begin{split}
        \frac{\partial}{\partial \alpha} KL(\psi_1 + \alpha(\phi - \psi_1)||\psi_1) &= \frac{\partial}{\partial \alpha} \left[\int (\psi_1 + \alpha(\phi-\psi_1) ) \log \left(1 + \alpha \left (\frac{\phi}{\psi_1} - 1 \right) \right) \right] \\
        &= \int (\phi-\psi_1) \log \left(1 + \alpha \left (\frac{\phi}{\psi_1} - 1 \right) \right),
    \end{split}
\end{equation}
where in the last step we have used the fact $\int \psi_1 = \int \phi = 1$. This directly shows $T_1=0$. Also, for any $\alpha \in [0,1]$, we have
\begin{equation}
    \begin{split}
        \frac{\partial^2}{\partial \alpha^2} KL(\psi_1 + \alpha(\phi - \psi_1)||\psi_1) &= \int \frac{(\phi-\psi_1)^2}{\psi_1 + \alpha(\phi-\psi_1)} \leq \chi^2(\psi_1||\phi) + \chi^2(\phi||\psi_1).
    \end{split}
\end{equation}
Now, by Cauchy--Schwartz inequality,
\begin{equation}
    \begin{split} 
        \chi^2(\psi_1||\phi)&=\int \frac{\left(\sum_{j=1}^k \beta_j (\phi_j-\phi)\right)^2}{\sum_{j=1}^k \beta_j \phi} \leq \int \sum_{j=1}^k \beta_j \frac{\left(\phi_j -\phi \right)^2}{\phi} = \sum_{j=1}^k \beta_j \chi^2(\phi_j||\phi), \\
        \chi^2(\phi||\psi_1)&=\int \frac{\left(\sum_{j=1}^k \beta_j \left(\phi -\phi_j \right) \right)^2}{\sum_{j=1}^k \beta_j \phi_j} \leq \int \sum_{j=1}^k \beta_j \frac{\left(\phi -\phi_j \right)^2}{\phi_j} = \sum_{j=1}^k \beta_j \chi^2(\phi||\phi_j).
    \end{split}
\end{equation}
Adding the above equations and then using (37), we have $T_2 \leq \sum_{j=1}^k \beta_j \left( \chi^2(\phi||\phi_j) + \chi^2(\phi_j||\phi) \right)$. Now from (35) and statement following (36), we can conclude that $\mathcal{D}_n(\psi_2| |\psi_1) \leq \frac{\alpha^2}{2}\sum_{j=1}^k \beta_j \left( \chi^2(\phi||\phi_j) + \chi^2(\phi_j||\phi) \right)$. This gives
\begin{equation}
    \mathcal{C}_n \leq \sup \Bigg\{\sum_{j=1}^k \beta_j \left( \chi^2(\phi||\phi_j) + \chi^2(\phi_j||\phi) \right) : \phi, \phi_1 \dots \phi_k \in \Gamma_n, \boldsymbol{\beta} \in \Delta^k \Bigg\},
\end{equation}
where this upper bound is a function of only $M,c_0$ and $\sigma_n$, appearing in the definition of $\Gamma_n$. We have reduced the upper bound calculation to that of $\chi^2$ divergence of single Gaussians, so we can now apply lemma 3. For any two members of $\Gamma_n$, say $\phi_2$ and $\phi_1$ , we get
\begin{equation}
    \chi^2\left(\phi_2||\phi_1\right) \leq (2-c_0)^{-d/2} \exp \left( \frac{2 M^2}{(2-c_0)\sigma_n} \right).
\end{equation}
Identical bound holds if we swap $\phi_1$ and $\phi_2$. Now this translates to
\begin{equation}
    \mathcal{C}_n \leq 2 (2-c_0)^{-d/2} \exp \left( \frac{2 M^2}{(2-c_0)\sigma_n} \right).
\end{equation}

\section*{PROOF OF LEMMA 3}

We first define the $\chi^2$ divergence between densities, which is a discrepancy measure comparable to Kullback--Leibler divergence and stronger than it. Refer to \cite{van2014renyi} for more details. \newline

\textbf{Definition 5:} For densities $\phi_1, \phi_2$ on $\mathbb{R}^d$, the $\chi^2$ divergence of $\phi_2$ from $\phi_1$ is defined as
\begin{equation}
    \chi^2(\phi_2| |\phi_1) = -1 + \int \frac{\phi_2^2}{\phi_1}.
\end{equation}
Choosing $\phi_i = \mathrm{N} \left(\mu_i,\sigma_i^2 \mathbf{I}_d\right), i=1, 2$, we have
\begin{equation}
    \frac{\phi_2^2}{\phi_1}(y) = \left( \sqrt{2 \pi} \frac{\sigma_2^2}{\sigma_1} \right)^{-d}  \exp \left[ \left( -\frac{1}{2} \right) \left(\frac{2\|y-\mu_2\|_2^2}{\sigma_2^2} - \frac{\|y-\mu_1\|_2^2}{\sigma_1^2} \right) \right].
\end{equation}
Let the term inside the exponent in (26) be $-B(y)/2$. We re-write
\begin{equation}
    B(y) = \left( \frac{2\sigma_1^2 - \sigma_2^2}{\sigma_1^2 \sigma_2^2} \right) \left\| y - \frac{\frac{2\mu_1}{\sigma_1^2} - \frac{\mu_2}{\sigma_2^2}}{\frac{2\sigma_1^2 - \sigma_2^2}{\sigma_1^2 \sigma_2^2}} \right\|^2_2 - \frac{2 \left\| \mu_2 - \mu_1 \right\|_2^2}{2\sigma_1^2 - \sigma_2^2},
\end{equation}
which shows
\begin{equation}
    \int_{y \in \mathbb{R}^d} \left( \sqrt{2 \pi} \right)^{-d}  \exp \left[ \left( -\frac{1}{2} \right) \left(\frac{2\|y-\mu_2\|_2^2}{\sigma_2^2} - \frac{\|y-\mu_1\|_2^2}{\sigma_1^2} \right) \right] dy = \exp \left( \frac{\left\| \mu_2 - \mu_1 \right\|_2^2}{2\sigma_1^2 - \sigma_2^2} \right) \left( \frac{\sqrt{2\sigma_1^2 - \sigma_2^2}}{\sigma_1 \sigma_2} \right)^{-d}.
\end{equation}
Combining (26) and (28), and then using (25), we get lemma 3.

\section*{AUXILIARY RESULTS}
In this section we provide proofs of the results $(4)-(7)$ in the main article. Since (5) follows directly from Bernstein-von-Mises theorem and its equivalence with (6) follows from
\begin{equation}
    d^2_{H} \leq d_{TV} \leq \sqrt{2} d_{H},
\end{equation}
we only focus on \newline

\textbf{Proposition 1:}
\begin{equation}
    \begin{split}
        KL \left(N\left( \theta_0, n^{-1}\Sigma \right)| |N\left( \mu_n, \Sigma_n \right) \right) & \rightsquigarrow \frac{1}{2} \chi^2_d, \\
        KL \left(N\left( \overline{X}_n, n^{-1}\Sigma \right)| |N\left( \mu_n, \Sigma_n \right) \right) & \rightarrow 0 \quad \mathrm{a.s.},
    \end{split}
\end{equation}
where '$\rightsquigarrow$' denotes weak convergence and a.s stands for almost sure validity with respect to the data generating distribution.

\section*{PROOF OF PROPOSITION 1}
We start with noting that
\begin{equation}
    \mu_n = \frac{n \overline{X}_n + \mu_0}{n + 1}, \quad \Sigma_n^{-1} = n \Sigma^{-1} + \Sigma_0^{-1}, \quad v_{1,n} := \mu_n - \theta_0, \quad w_{1,n} := \mu_n - \overline{X}_n.
\end{equation}
Observe the difference between $v_{1,n}$ and $w_{1,n}$ by noting
\begin{equation}
    \begin{split}
        \sqrt{n} v_{1,n} &= \frac{ \sqrt{n} \left( \overline{X}_n - \theta_0 \right)}{1 + \frac{1}{n}} + \frac{\sqrt{n}}{n + 1} \left( \mu_0 - \theta_0 \right) =: v_{2,n} + v_{3,n}, \\ 
        \sqrt{n} w_{1,n} &= \frac{\sqrt{n}}{n + 1} \left( \mu_0 - \overline{X}_n \right) =: w_{2,n}.
    \end{split}
\end{equation}
We now have from (48) and (49)
\begin{equation}
    \begin{split}
        v_{1,n}^T \Sigma_n^{-1} v_{1,n} &= \left( v_{2,n} + v_{3,n} \right)^T \left[ \Sigma^{-1} + n^{-1} \Sigma_0^{-1} \right] \left( v_{2,n} + v_{3,n} \right) \\
        &= \left( v_{2,n} + v_{3,n} \right)^T \Sigma^{-1} \left( v_{2,n} + v_{3,n} \right) + n^{-1} \left( v_{2,n} + v_{3,n} \right)^T \Sigma_0^{-1} \left( v_{2,n} + v_{3,n} \right) =: E_{1,n} + n^{-1} E_{2,n}.
    \end{split}
\end{equation}
Let us deal with $E_{2,n}$ first. Breaking down further, we see
\begin{equation}
    E_{2,n} = v_{2,n}^T \Sigma_0^{-1} v_{2,n} + 2 v_{2,n}^T \Sigma_0^{-1} v_{3,n} + v_{3,n}^T \Sigma_0^{-1} v_{3,n}.
\end{equation}
The last term in the right-hand-side of (51) is non-stochastic, and $\Sigma_0, \mu_0$ are free of $n$. Hence, we get
\begin{equation}
    v_{3,n}^T \Sigma_0^{-1} v_{3,n} = \frac{n}{(n + 1)^2} \left( \mu_0 - \theta_0 \right)^T \Sigma_0^{-1} \left( \mu_0 - \theta_0 \right) \to 0.
\end{equation}
The second term in the right-hand-side of (51) satisfies
\begin{equation}
    v_{2,n}^T \Sigma_0^{-1} v_{3,n} = \left( \frac{n}{n + 1} \right)^2 \left[ \left( \mu_0 - \theta_0 \right)^T \Sigma_0^{-1} \right] \left( \overline{X}_n - \theta_0 \right) \to 0 \quad a.s.,
\end{equation}
using Strong Law of Large Numbers(SLLN). Now for the first term on the right-hand-side of (51), we have
\begin{equation}
    n^{-1} v_{2,n}^T \Sigma_0^{-1} v_{2,n} = \left( \frac{n}{n + 1} \right)^2 \left( \overline{X}_n - \theta_0 \right)^T \Sigma_0^{-1} \left( \overline{X}_n - \theta_0 \right) \to 0 \quad a.s.,
\end{equation}
where we have again used SLLN. Putting together (51)--(54), we get that
\begin{equation}
    n^{-1} E_{2,n} \to 0 \quad a.s.
\end{equation}
We now deal with $E_{1,n}$ in the right-hand-side of (50), observing
\begin{equation}
    E_{1,n} = v_{2,n}^T \Sigma^{-1} v_{2,n} + 2 v_{2,n}^T \Sigma^{-1} v_{3,n} + v_{3,n}^T \Sigma^{-1} v_{3,n}.
\end{equation}
Similar to (52)--(53), we have
\begin{equation}
    v_{3,n}^T \Sigma^{-1} v_{3,n} \to 0, \quad v_{2,n}^T \Sigma^{-1} v_{3,n} \to 0 \; a.s.,
\end{equation}
while the first term on the right-hand-side of (56) satisfies
\begin{equation}
    v_{2,n}^T \Sigma^{-1} v_{2,n} = \left( \frac{n}{n + 1} \right)^2 \left\| \Sigma_0^{-1/2} \sqrt{n} \left( \overline{X}_n - \theta_0 \right) \right\|_2^2 \rightsquigarrow \chi_d^2 .
\end{equation}
by an application of Slutsky's theorem, as the random vector within norm signs in (58) follows a standard Gaussian distribution in $d$-dimensions under the true vector $\theta_0$. Another application of Slutsky's theorem allows us to get from (50)
\begin{equation}
    v_{1,n}^T \Sigma_n^{-1} v_{1,n} \rightsquigarrow \chi_d^2.
\end{equation}
We next turn our attention to
\begin{equation}
    w_{1,n}^T \Sigma_n^{-1} w_{1,n} = w_{2,n}^T \left[ \Sigma^{-1} + n^{-1} \Sigma_0^{-1} \right] w_{2,n} = w_{2,n}^T \Sigma^{-1} w_{2,n} + n^{-1} w_{2,n}^T \Sigma_0^{-1} w_{2,n} := F_{1,n} + n^{-1} F_{2,n},
\end{equation}
to get
\begin{equation}
    \begin{split}
        F_{1,n} &= \frac{n}{(n + 1)^2} \left( \overline{X}_n - \mu_0 \right)^T \Sigma^{-1} \left( \overline{X}_n - \mu_0 \right) \to 0 \quad a.s., \\
        F_{2,n} &= \frac{n}{(n + 1)^2} \left( \overline{X}_n - \mu_0 \right)^T \Sigma_0^{-1} \left( \overline{X}_n - \mu_0 \right) \to 0 \quad a.s.
    \end{split}
\end{equation}
Note that the limit is driven by the preceding factor of $n/(n + 1)^2$ rather than SLLN. We thus have
\begin{equation}
    w_{1,n}^T \Sigma_n^{-1} w_{1,n} \to 0 \quad a.s.
\end{equation}
We are now prepared for the final part of the proof where we use the well-known formula for KL divergence between Gaussians to get
\begin{equation}
    \begin{split}
        KL \left(N\left( \theta_0, n^{-1}\Sigma \right)| |N\left( \mu_n, \Sigma_n \right) \right) &= \frac{1}{2} \left[ \mathrm{trace} \left( n^{-1} \Sigma_n^{-1} \Sigma \right) - d - \log \det \left( n^{-1} \Sigma_n^{-1} \Sigma \right) + v_{1,n}^T \Sigma_n^{-1} v_{1,n} \right] \\
        KL \left(N\left( \overline{X}_n, n^{-1}\Sigma \right)| |N\left( \mu_n, \Sigma_n \right) \right) &= \frac{1}{2} \left[ \mathrm{trace} \left( n^{-1} \Sigma_n^{-1} \Sigma \right) - d - \log \det \left( n^{-1} \Sigma_n^{-1} \Sigma \right) + w_{1,n}^T \Sigma_n^{-1} w_{1,n} \right]
    \end{split}
\end{equation}
Most of the expressions are common for both the equations in (63), and we evaluate them term by term. We have
\begin{equation}
    \begin{split}
        &\mathrm{trace} \left( n^{-1} \Sigma_n^{-1} \Sigma \right) - d = \mathrm{trace} \left( I_d + n^{-1} \Sigma_0^{-1} \Sigma \right) - d \to 0, \\
        &\log \det \left( n^{-1} \Sigma_n^{-1} \Sigma \right) = \log \det \left( I_d + n^{-1} \Sigma_0^{-1} \Sigma \right) \to 0.
    \end{split}
\end{equation}
Proposition 1 now follows by combining (59) and (62)--(64), along with an application of Slutsky's theorem.

\bibliography{refs}
\bibliographystyle{plainnat}

\end{document}